# Multi-lingual Geoparsing based on Machine Translation

Xu Chen[1,2*], Han Zhang, Judith Gelernter

State Key Laboratory of Software Engineering, Wuhan University, Wuhan, China

Language Technologies Institute, School of Computer Science, Carnegie Mellon University, Pittsburgh, USA

Email: xuchen@whu.edu.cn hanz_cmu@yahoo.com gelern@cs.cmu.edu

*Abstract*—**Our method for multi-lingual geoparsing uses monolingual tools and resources along with machine translation and alignment to return location words in many languages. Not only does our method save the time and cost of developing geoparsers for each language separately, but also it allows the possibility of a wide range of language capabilities within a single interface. We evaluated our method in our LanguageBridge prototype on location named entities using newswire, broadcast news and telephone conversations in English, Arabic and Chinese data from the Linguistic Data Consortium (LDC). Our results for geoparsing Chinese and Arabic text using our multi-lingual geoparsing method are comparable to our results for geoparsing English text with our English tools. Furthermore, experiments using our machine translation approach results in accuracy comparable to results from the same data that was translated manually.**

*Keywords: Named Entity Recognition (NER), Location, Geoparser, Machine Translation, Word Alignment*

## I. INTRODUCTION

The Internet of Things embraces the web 2.0 of web pages and social media, but adds devices their services. More people will speak about devices and services in more languages, making inter-language communication even more vital. This research will help.

Our research concerns how to find named entities, and location names in particular, in different languages. Some location mapping takes given locations in latitude, longitude coordinates. Our research concerns the difficult problem of when we need to extract location words, and associate latitude and longitude coordinates with the centroid of that named place.

It has been found that named entities in many languages withstand machine translation such that, when machine translated into English, we can use English Named Entity Extraction tools. This is the basis of our method. However, we have concentrated on alignment between the source language and English, and have performed experiments to demonstrate our method's validity.

The beauty of our method is its cost-effectiveness. Due to the wide reach of machine translation tools (third-party software in our workflow), we are able to find named entities in a range of language, some of which are poor in entity extraction tools. Thus, using our method in our LanguageBridge system, a wide range of languages can be geo-parsed using English tools.

*Research questions*

o   How can we improve word alignment so as to improve the accuracy of the output?

o   How does our precision and recall in geoparsing without machine translation (in English) compare generally with precision and recall in geoparsing that relies on machine translation (in Chinese and Arabic)?

o   To what extent does translation quality influence the geoparsing result?

*Experiments to solve research questions*

For data, we used files in Chinese and Arabic that included manual translations in English, as well as English files.

For running experiments, we used our LanguageBridge system based on Machine Translation, and we compared to the BasisTech Rosette system that has a Chinese component that finds locations in Chinese (without Machine Translation). We showed that our method based on Machine Translation is almost as accurate as geoparsing in the source language.

We also used our method with our Geolocator geoparser and Yahoo's Geomaker geoparser – on human translations as well as machine translations. We showed that our method works as well using either geoparser, and more surprisingly, considering translation quality, that the results are almost as high when geoparsing a machine translation as a manual translation.

Section §2 describes related work, and §3 describes the methodology for our multi-lingual geoparser, LanguageBridge, and details the sub-processes required for each step of our implementation. Data for the experiment data is described in §4. Evaluation experiments for the LanguageBridge appear in §5. The paper concludes in §6 with potential research directions and a summary of our contributions.

## II. RELATED WORK

Our method is based on previous research that finding locations in Spanish tweets with a geoparser trained for Spanish was less accurate than geoparsing an English translation of the same Spanish tweets with a geoparser trained for English (Gelernter and Zhang, 2013). Similar results were found when using machine translation and English tools to find named entities in source texts in Swahili and Arabic (Shah et al, 2011). In fact, statistical machine translation is often used for cross-language information retrieval (Nikoulina and Clinchant, 2013).

Bilingual texts, also called parallel corpora, have been used to strengthen monolingual Named Entity Recognition algorithms (Che et al, 2013) (Wang et al, 2013), and create named entity annotations (Ehrmann et al, 2011). Our objective, by contrast, is to find location named entities in texts in many languages immediately by adding machine translation tools to an already strong Named Entity Recognizer—without additional training when possible.

The Cross-Language Retrieval Forum, CLEF, ran geo-tracks in 2005 (a pilot year) 2006, 2007 and 2008, in order to test the ability of a system to find location information in multiple languages (Mandl et al, 2008). The CLEF experiments differ from ours in that participating systems were expected to answer questions regarding location to express geographical relationships (proximity, inclusion and exclusion), rather than just to identify locations, as ours does.

Further, our method using Google or Microsoft Translator permits a wide range of language capabilities with very little additional coding required. Thus, our method allows us to find location expressions in dozens of languages due to the translation range of Google and Microsoft. This is in comparison to the GeoCLEF experiments in 2008 that were in European languages only, and Rosette NER from BasisTech can find locations (and in fact, standard named entities) in 16 languages at the time of this writing.[1] However, our use of our own Geoparser that has Stanford NER, our own CRF-trained classifier and some heuristics, makes our algorithm more robust (Gelernter and Zhang, 2013).

## III. METHODOLOGY

### A. Multi-Lingual Geoparsing with Language Bridge

Our prototype system called LanguageBridge was implemented using a component-based approach that includes our own geoparser, a Machine Translation component (whether Google or Microsoft), and word alignment (whether from Google or Microsoft) along with our alignment adjustment scripts. A developer might substitute another component for ours to alter the processing result.

Figure 1 illustrates the procedure of multi-lingual geo-parsing. Word alignment is created between the input language and English when the machine translation algorithm is run. We have added scripts to improve the alignments, as discussed below, as well as a hash map to store the alignment.

Next, our English geoparser finds locations in the English translation. The multi-lingual geoparser uses the alignment information to match the found locations with those in the original language.

The last step is for the location words identified in the English translation to be displayed both in English and in the original language. There is an option to display also the latitude and longitude coordinates for location.

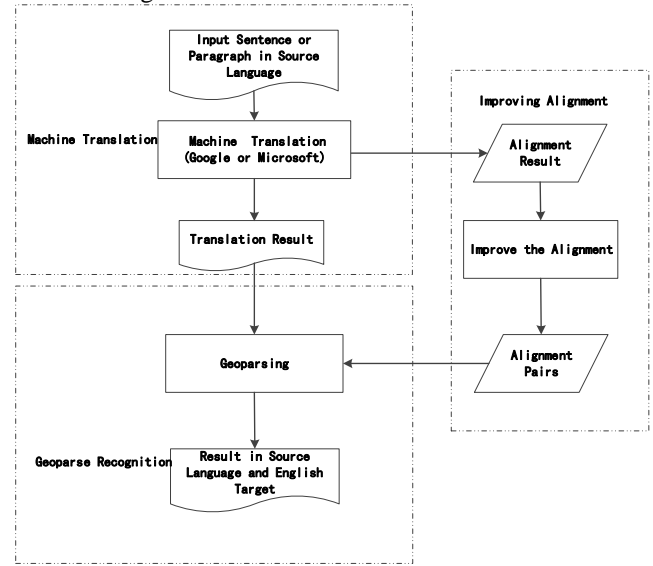

**Fig. 1** Procedure for our multi-lingual geoparser, LanguageBridge

### B. Word alignment adjustments

Our multi-lingual geoparser is based on machine translation (from either Microsoft or Google), and word alignment algorithms.

*Adjustments for locations found in the source language.*

The word alignment from machine translation often loses words, or combines function words with toponyms so as to complicate the geoparsing. Hence, we geoparse the coarse alignment before re-aligning with the source language.

1. Some non-English tokens bundle a preposition with a named entity. For example, in Microsoft alignment in the Russian language, "Крыма = of Crimea." Our algorithm must separate preposition from location in order to identify the location.

2. Synonymous Chinese words align with the same word in English. For example, even though "塞尔维亚共和国" translates as "Republic of Serbia", and "塞 尔 维 亚" translates as "Serbia", when they appear in the same sentence, both "塞尔维亚共和国" and "塞尔维亚" align with "Serbia," and our algorithm chooses only one in the pair for the hash map.

3. Some words may lose alignment information during machine translation. However, if the word occurs two or more times in the same sentence, we can supplement the lost alignment information based on the same word in other places of sentence. For example, if we find "اسرائیل" aligns with "Israel," then the algorithm will use "Israel" in other parts of the translated sentence.

---

**Algorithm1** alignment information improvement

**Input**: Alignment String S(sourceNumSeq₁-enNumSeq₁,…,sourceNumSeqᵢ-enNumSeqᵢ,…, sourceNumSeqₙ-enNumSeqₙ).

**Output**: HashMap pairs.

Scanner alignmentScanner = new Scanner(S);

paris = new HashMap();

While (alignmentScanner.hasnext){

    String wordAlignment = alignmentScanner.next();

    String newValue = FindWᵢ (wordAlignment.getSourceNumSeqᵢ());

    newValue.removePreopostion();

    String Key = FindW'ᵢ (wordAlignment.getEnNumSeqᵢ());

    String oldValue = pairs.getValue (Key);

    If (oldValue==null||oldValue.equals(newValue))

    pairs.put (Key, newValue);

    else if

    String Value = Compare (oldValue, newVaule);

    paris.put (key,Value);   }

return pairs;

---

We propose Algorithm1 to improve the alignment information from machine translation. The input is the Alignment String and output is a hash map which stores the pairs of word or phrases for the source language and English.

Errors are caused when the oldValue and newValue differ but share the same key. The algorithm compares the values in the sentence to see whether one is one a subset of the other. For example, both "塞尔维亚共和国" Republic of Serbia and "塞尔维亚" Serbia are in the original sentence, with the second as a subset of the first. We created a rule to output the longer version from the original. Alternatively, we could retain the alignment positions of both versions of the toponym from the original text and use different keys to store in the hash map, but this would make it take more time to process the alignment and output the result.

*Adjustments for locations found in the Machine Translation into English*

When we find the locations that are output by our English geoparser, we should get the location words in source language too. But sometimes we cannot find the alignment information directly.

1. Adjacent English words align with the same word in Chinese. For example, "United" maps to "美国," and also "States" maps to "美国".

2. Alignment information is not available. The alignment cannot find a phrase in Chinese to match "Carnegie Mellon University". Therefore, it finds the words one by one rather than a phrase, and then combines them to give a result.

---

**Algorithm2** find source toponym

**Input** : Toponym Set in English, enToponym(t₁, …,tᵢ,…tₙ)

**Output** : Toponym Pairs of Chinese and English toponymPair(t₁---t'₁, …, tᵢ---t'ᵢ, …, tₙ---t'ₙ).

    for (topnymEntity tᵢ: enToponym){

    String key = tᵢ;

    String t'ᵢ = getOriginalWords(key);

    if (t'ᵢ == null){

    StringTokenizer toponymTok = new StringTokenizer(key);

        while (toponymTok.hasMoreTokens){

        String temporary = toponymTok.nextToken();

        if ((temporary != null)&&!temporary.equals(t'ᵢ )){

    t'ᵢ += temporary;   }   }   }

    addToponymPair(tᵢ,t'ᵢ)

    }

    return toponymPair;

---

Our implementation takes all of these errors into account to improve source − target alignment of multi-word location expressions.

Algorithm2 fixes errors after the location words in English are output from the geoparser. For the example of "Great Britain," the alignment algorithm cannot identify Great Britain as a phrase, so the hash map stores "Great" and

"Britain" as separate keys. Algorithm2 restores the values of the two keys in the output.

### C. Machine Translation and Alignment based on Google

The Google Translate API v2 [2] does not provide alignment information. A Google employee advised us to approximate alignment information by doing HTML translation, where we propagate HTML tags from the source to the target. But he warned that sometimes doing this can affect translation quality, as the system tries to preserve the HTML formatting, which might confuse location expressions.[3]

We used html tags to separate each word in the source sentence. For a source language in Arabic, the markup looks like this: <font font="1">إسرائيل</font>

We assemble these words into an html file, which becomes our input for Google Translate. Then we get the translation result from Google. We align the word or phrase based on the <html> tag number. As shown in Table 1, we can get the Arabic-English pairs for the toponym, e.g.Israel - إسرائيل. Finally we store the alignment in a hash like this: {Israel=إسرائيل, in=في, I=أنا, live=أعيش}.

**Table 1.** Translation of the Arabic أنا أعيش في إسرائيل based on Google Translation

| Arabic | English |
|---|---|
| <html lang="en-x-mtfrom-ar"> | <html lang="en-x-mtfrom-ar"> |
| <head></head> | <head></head> |
| <body> | <body> |
| <doctype html=""> | <doctype html=""> |
| <title></title> | <title></title> |
| <font font="1">أنا</font> | <font font="1">I</font> |
| <font font="2">أعيش</font> | <font font="2">live</font> |
| <font font="3">في</font> | <font font="3">in</font> |
| <font font="4">إسرائيل</font> | <font font="4">Israel</font> |
| </doctype> | </doctype> |
| </body> | </body> |
| </html> | </html> |

### D. Machine Translation and alignment based on Microsoft

Microsoft's online statistical translation service, is Microsoft Translator.[4] In our experiment, we use the SOAP API from Microsoft that provides alignment information as well as the translation. The Simplified Chinese sentences below have been geoparsed and aligned based on Microsoft.

---

**Source**: 美国在加勒比海和太平洋还拥有多处领土和岛屿地区

**Translation from Microsoft:**[The] United States [is] in the Caribbean and the Pacific, [and] also has a number of territories and insular areas

**Alignment information:**0:1-0:5 0:1-7:12 2:2-14:15 3:5-21:30 6:6-21:30 7:7-31:33 8:10-39:45 11:11-48:51 12:13-53:55 14:14-59:64 16:17-69:79 18:18-81:83 19:20-85:91 21:22-93:97

**HashMap that we create from the alignment:**{territories=领土, Pacific=太平洋, Caribbean=加勒比海, areas=地区, number=多, insular=岛屿, also=还, in=在, has=拥有, States=美国, United=美国, and=和}

**English Geoparser:** Caribbean, Pacific, United States

**LanguageBridge output:** Caribbean----加勒比海, Pacific----太平洋,United States----美国

---

The alignment between any two languages is straightforward when it consists of one source token to one target token, or one source token to many target tokens, because a single concept may be expressed by multiple tokens. In the example above, aligning to English Caribbean and Pacific are likely correct. Algorithm2 fixed the mechanical error that arises in the translation when two or more tokens aligns with two or more tokens. In the example above, the error is that the United States is repeated twice, with 美国美国.

## IV. DATA

In order to test our multi-lingual geoparsing method, we used Automatic Content Extraction (ACE) 2005 Multilingual Training Corpus LDC2005E18. This includes three separate data sets for English, Chinese and Arabic. We relied on the ACE annotations for experiments reported here.

The data corpus includes three tags which we consider locations: GPE, LOC, and NAM. GPE stands for geopolitical entities, LOC for location, and NAM[5] for some references to names that include location. We use texts in the corpus taken from Newswire, Broadcast News, Broadcast Conversation, Conversational Telephone Speech, and we randomly chose about 100 files for each language for testing.

**Table 2.** Testing data from about 100 files from each language of ACE 2005Multilingual LDC2005E18

| | number of words | number of locations | number of unique locations |
|---|---|---|---|
| *Chinese* | 33349 | 912 | 238 |

| | number of words | number of locations | number of unique locations |
|---|---|---|---|
| *Arabic* | 20087 | 1435 | 348 |
| *English* | 31255 | 851 | 243 |

We selected this number of files to roughly balance the number of unique locations among languages. In Table 2, our count of the number of words, the number of locations and unique locations are based on the annotations provided in the ACE Multi 2005 data set.

We selected two sets of LDC parallel corpora with Newswire text, 2012T16 (Chinese-English) and 2014T05 (Arabic-English), because they include high-quality English translations by bilingual speakers for Chinese and Arabic that we could use to compare with the machine translations. In Table 3, we created our own annotations in order to count the locations and unique locations.

**Table 3.** Testing data from about 50 files for each language from the Parallel Corpora LDC2012T16 and LDC2014T05

| | number of words | number of locations | number of unique locations |
|---|---|---|---|
| *Chinese* | 20357 | 469 | 113 |
| *Arabic* | 13422 | 850 | 182 |

## V. EVALUATION OF OUR LANGUAGEBRIDGE PROTOTYPE FOR MULTI-LINGUAL GEOPARSING

We tested the accuracy of our method with three experiments.

Exp_1a). How does finding locations in the same language as the tool (native English with an English tool) compare to finding locations in machine translation into English with an English tool.

**Table 4.** Geoparsing for location words from LDC2005E18: Precision, Recall and F1 for Chinese, Arabic and English

| | Precision | Recall | F1 |
|---|---|---|---|
| **Chinese** | 0.821 | 0.737 | 0.777 |
| **Arabic** | 0.781 | 0.784 | 0.782 |
| **English** | 0.887 | 0.826 | 0.855 |

*Experimental procedure:* We geoparse the English directly with our own geoparser (Gelernter and Zhang, 2013), and also we are using our multi-lingual geoparsing methods that include machine translation (here, with Microsoft and Google Translator) to output locations in Chinese and Arabic. We selected files for testing from ACE 2005Multilingual LDC2005E18.

*Experiment results:* Precision in all three languages suffers due to the geo/non-geo disambiguation problem of non-geographic names (example: Jordan as a man's name) being mistaken for a toponym (Jordan, the country). Results in Table 4 show that the overall F1 of Chinese is close to Arabic, and both of them are comparable to the output of English.

*Result analysis:* Note that the precision in Chinese is higher than that in Arabic. Some of the variability can be explained by the fact that the Chinese→ English path is easier than the Arabic→English path. It has been said that up to 75% of this variability can be explained by factors such as the amount of word reordering necessary, and the historical relatedness of the two languages (Birch et al, 2008).

Exp_1b). How does finding locations in the same language as the tool (native Chinese with a Chinese tool, Rosette NER) compare to finding locations in machine translation into English with an English tool (Language Bridge).

*Experimental procedure:* We selected 106 test files in Chinese from ACE2005 Multilingual LDC2005E18. These files were run both in the Chinese version of the Rosette NER by BasisTech, and in our LanguageBridge. The Rosette NER finds other named entities, but we scored only for location.

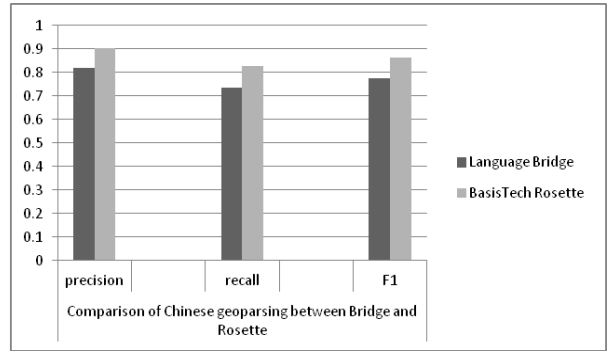

Fig. 2 Comparison of geoparsing from a translation with the Carnegie Mellon Language Bridge vs geoparsing in the original language with BasisTech Rosette NER on 106 test files in Chinese from ACE2005 Multilingual LDC2005E18.

*Experimental results:* Figure 2 demonstrates that somewhat higher results were achieved in using the BasisTech software than in using our English-based tool.

*Result analysis*: As demonstrated by Figure 2, quality geoparsing in the original language has the potential to achieve better results than in parsing via machine translation. Nevertheless, it has been found that named entities can be identified with success via Machine Translation for Arabic and Swahili (Shah et al, 2011), and also for Spanish (Gelernter and Zhang, 2013)—and that, as shown in Figure

2, the differential between the in-Chinese parsing and the cross-language parsing is not high.

*Significance.* Significant for our argument is that the comparison cannot be performed in many languages (Rosette NER presently supports 16 languages only). Our system, by contrast, handles dozens of languages owing to the range of Google Translate and Microsoft Translate. This demonstrates the wide significance of the "black box" method using machine translation for location detection.

Exp_2. Given the results of Exp_1 that geoparsing translations (with Named Entity Recognizers) achieves solid results, to what extent does the translation quality matter?

*Experimental procedure:* We tested our LanguageBridge multi-lingual geoparser on data sets in which the same text is provided in two languages, and in both manual and machine translation: Arabic and English (parallel corpus LDC2014T05) and Chinese and English (parallel corpus LDC2012T16). We randomly selected 50 files from each data set and used the manual translations to annotate the original. Then we used the Microsoft Machine translation algorithm with LanguageBridge to find locations in Chinese, and the Google Machine translation algorithm with LanguageBridge to find locations in the Arabic.

**Table 5** Geoparsing for location words: Precision, Recall and F1 found through
Machine translation vs. manual translation

|  | Chinese | | Arabic | |
|---|---|---|---|---|
|  | *Machine translation* | *Manual translation* | *Machine translation* | *Manual translation* |
| *Precision* | 0.796 | 0.810 | 0.708 | 0.758 |
| *Recall* | 0.776 | 0.783 | 0.923 | 0.942 |
| *F1* | 0.786 | 0.796 | 0.801 | 0.840 |

*Experimental results:* Table 5 shows that the precision and recall of machine translation results approaches precision and recall achieved by manual translation. According to Pearson's Product-Moment Correlation, we found no statistically significant difference in the geoparsing precision and recall between manual and machine translation for Chinese, and also no statistically significant difference in geoparsing precision and recall between manual and machine translations for Arabic.

*Result analysis:* Why does the translation quality for finding locations in Chinese and Arabic text seem insignificant? The Named Entity Recognition that is the basis for finding locations does not rest on subtle text understanding. Instead, Named Entity Recognition relies upon correct recognition of part of speech of words, some location-indicating phrases, and location-word matches with a gazetteer, all of which can be accomplished adequately from a good machine translation.

Exp_3. How robust is our cross-lingual geoparsing method?

*Experimental procedure.* Parallel corpora are used for statistical machine translation and other procedures for Natural Language Processing. The Linguistic Data Consortium includes parallel corpora (2014T05—Arabic/English) and (2012T16—Chinese/English) that include both machine and manual translations. We selected 50 files at random from each corpus, and annotated the locations found in those files. The Arabic set had 182 unique toponyms, and the Chinese set had 113 (see Table 3). We sent both manual and machine translations from each language through our own Geoparser, and through the Yahoo GeoMaker, (1) to compare parsing tools. We were interested also (2) in comparing relative accuracy between Chinese and Arabic, and (3) in comparing relative accuracy with different translation quality.

*Experimental results.* Three results comes from this experiment. (1) Tools: The first is that our geoparser is comparable to Yahoo GeoMaker in precision and recall, both for the Chinese and for the Arabic data set. The GeoMaker outperforms our Geolocator in precision for both languages, but the Geolocator dominates the GeoMaker in recall for Arabic, making it outperform the GeoMaker in Arabic overall. (2) Languages: The files were chosen randomly from Arabic and Chinese. There are more unique toponyms in the Arabic data set than in the Chinese. Proportionally, however, the ability to find locations accurately in both languages is comparable for our Geolocator (whereas the GeoMaker performed better in Arabic than in Chinese).(3) Translation quality: Results demonstrate finally that the quality of the translation matters little in results – that the locations found in the machine translation approximate those in the manual translation for both languages. The Geomaker even found more locations accurately based on Chinese machine translation than on the manual.

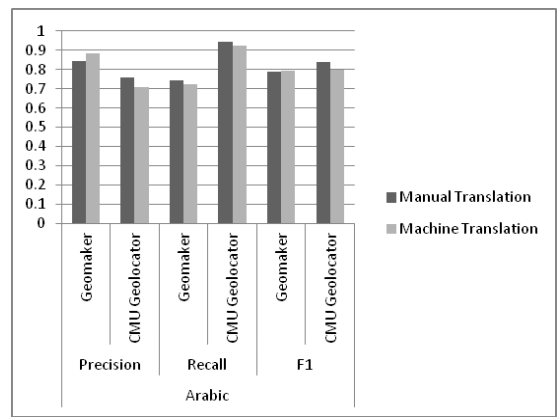

**Fig. 3 Our geoparser compared to GeoMaker on 50 files in Arabic and English (parallel corpus LDC2014T05)**

*Result analysis:* The overall results of Figure 3 and 4 demonstrate the effectiveness of our method in using machine translations of text with English geoparsing tools.

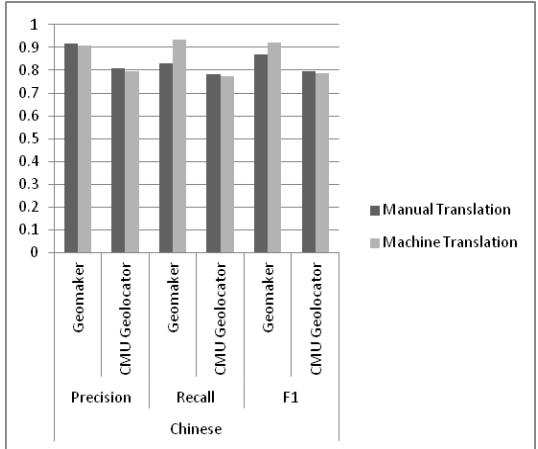

**Fig. 4 Our geoparser compared to GeoMaker on 50 files in Chinese and English (parallel corpusLDC2012T16)**

## VI. CONCLUSION

We propose a cost-efficient method to build a multi-lingual geoparser based on machine translation and word alignment adjustment. We demonstrated the viability of our system by running our multi-lingual geoparser over Chinese and Arabic text, as well as over English text. The experiment confirmed that results from geoparsing Arabic and Chinese were of comparable accuracy to results in English.

We found that geoparsing the machine translation into English from Chinese and Arabic yields results comparable to geoparsing the high quality manual translations into English from Chinese and Arabic.

Validating our method, we found that results with English geoparsing tools other than ours were comparable to results with our own Language Bridge. Notwithstanding variability in Machine Translation results from every language, for a data set of similar size in a language which can be translated by Google or Microsoft, we expect that the location word output using our method will be fairly accurate.

## ACKNOWLEDGMENT

This work was supported by the grants of the National Natural Science Foundation of China (41201405), China Scholarship Council No. 201308420300.